\documentclass[10pt,twocolumn,letterpaper]{article}

\usepackage[pagenumbers]{cvpr} 

\usepackage{graphicx}
\usepackage{amsmath}
\usepackage{amssymb}
\usepackage{booktabs}
\usepackage[accsupp]{axessibility}

\usepackage[pagebackref,breaklinks,colorlinks]{hyperref}

\usepackage[capitalize]{cleveref}
\crefname{section}{Sec.}{Secs.}
\Crefname{section}{Section}{Sections}
\Crefname{table}{Table}{Tables}
\crefname{table}{Tab.}{Tabs.}

\def\cvprPaperID{3295} 
\def\confName{CVPR}
\def\confYear{2023}

\begin{document}

\title{Dynamic Graph Enhanced Contrastive Learning for Chest X-ray Report Generation}

\author{Mingjie Li$^{1}$\hspace{2mm}
Bingqian Lin$^2$\hspace{2mm}
Zicong Chen$^5$\hspace{2mm}
Haokun Lin$^2$ \hspace{2mm}
Xiaodan Liang$^{2,3,4}$ \hspace{2mm}
Xiaojun Chang$^1$\thanks{Corresponding author. \url{https://github.com/mlii0117/DCL}} \\
$^1$ReLER, AAII, University of Technology Sydney \\
$^2$School of ISE, Sun Yat-Sen University \\
$^3$Department of Computer Vision, Mohamed bin Zayed University of Artificial Intelligence \\
$^4$Peng Cheng National Lab~~$^5$The University of Hong Kong}

\maketitle

\begin{abstract}
Automatic radiology reporting has great clinical potential to relieve radiologists from heavy workloads and improve diagnosis interpretation. Recently, researchers have enhanced data-driven neural networks with medical knowledge graphs to eliminate the severe visual and textual bias in this task. The structures of such graphs are exploited by using the clinical dependencies formed by the disease topic tags via general knowledge and usually do not update during the training process. Consequently, the fixed graphs can not guarantee the most appropriate scope of knowledge and limit the effectiveness. To address the limitation, we propose a knowledge graph with \textbf{D}ynamic structure and nodes to facilitate medical report generation with \textbf{C}ontrastive \textbf{L}earning, named DCL. In detail, the fundamental structure of our graph is pre-constructed from general knowledge. Then we explore specific knowledge extracted from the retrieved reports to add additional nodes or redefine their relations in a bottom-up manner. Each image feature is integrated with its very own updated graph before being fed into the decoder module for report generation. Finally, this paper introduces Image-Report Contrastive and Image-Report Matching losses to better represent visual features and textual information. Evaluated on IU-Xray and MIMIC-CXR datasets, our DCL outperforms previous state-of-the-art models on these two benchmarks.
\end{abstract}
\vspace{-1cm}
\section{Introduction}
\label{sec:intro}

Recently, automatic report generation has received growing attentions from both machine learning and automatic medicine fields. It aims to generate semantically coherent and informative reports to describe the referring examination images, such as Chest X-Ray~\cite{iuxray,Johnson2019MIMICCXRAL}, Lung CT Scan~\cite{li2022auxiliary} or funds angiography~\cite{li2021ffa}. Such techniques have great clinical potential in relieving junior radiologists from heave workloads and reducing diagnosis errors by improving the interpretation~\cite{R2gen,liu2021exploring}. 

\begin{figure}[t]
  \centering
   \includegraphics[width=0.99\linewidth]{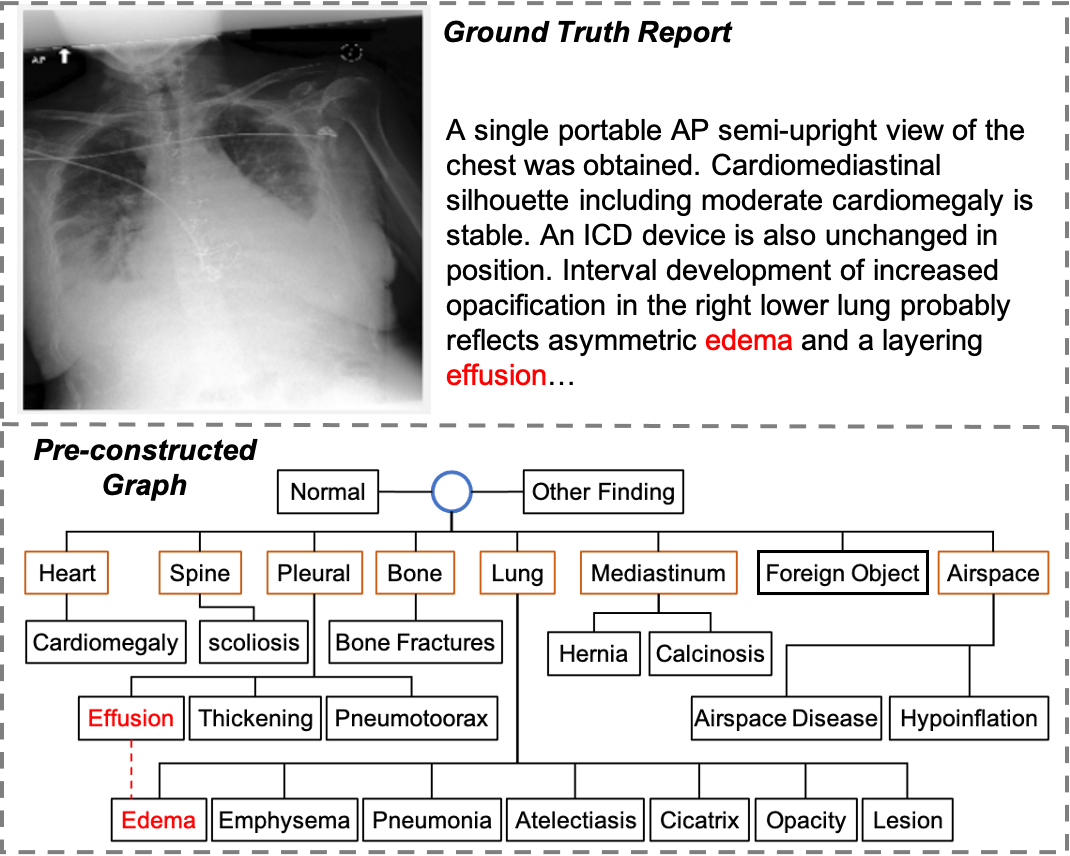}
   \caption{An illustration of one sample from MIMIC-CXR~\cite{Johnson2019MIMICCXRAL} and the pre-constructed graph in \cite{zhang2020radiology}, where the blue circle, orange boxes and black boxes refer to the global node, organ-level entities and key findings, respectively. The red dash line here represents the unconnected relation.}
   \label{fig:motivation}
   \vspace{-0.8cm}
\end{figure}

Witnessed the great progress in artificial intelligence, especially deep learning methods~\cite{resnet,vaswani2017attention,li2022video}, researchers have proposed various data-driven neural networks for radiology reporting and achieved promising performances in metrics that measure descriptive accuracy~\cite{R2gen,yang2022knowledge} and clinical correctness~\cite{endo2021retrieval,zhang2020radiology}. Compared with the similar task generic image captioning~\cite{hossain2019comprehensive}, the key challenges in medical report generation (MRG) task are the severe visual and textual data bias~\cite{Li2018HybridRR,li2022auxiliary}. On the one hand, medical images are highly similar to each other due to the imaging methods and human tissues themselves. However, abnormal regions or lesions that should acquire more attentions usually locate at a small part and lack detailed annotations in existing MRG benchmarks. On the other hand, sentences that describe normal regions are likely to appear repeatedly among each dataset which disables the model to describe specific crucial abnormalities. Two concepts have been proved effective in eliminating those bias. 

The first one is to integrate medical knowledge with MRG systems~\cite{li2022cross,zhang2020radiology,liu2021exploring,yang2022knowledge}. Zhang \etal~\cite{zhang2020radiology} constructed a universal graph comprised of a global node, 7 organs/tissues and 20 findings (normal or disease keywords). Disease keyword nodes linked to the same organ are connected to each other and the root in the graph. This graph can enhance the relationships among findings and emphasize the disease keywords. Thus, it is also adopted in the following works~\cite{liu2021exploring,yang2022knowledge}. However, this graph is built from general knowledge and may be inappropriate for some cases. As the shown report in \cref{fig:motivation}, it is observed that \textit{effusion} should be suggestive of \textit{edema}, however such relationship is not modelled in the graph. Furthermore, some nodes like `cicatrix' or `hypoinflation' only appear very few times in two MRG benchmarks~\cite{iuxray,Johnson2019MIMICCXRAL}. Therefore, it is necessary to update the scope of knowledge for each case; In addition to the medical knowledge, recent works~\cite{endo2021retrieval,liu2021contrastive,yan2021weakly,song2022cross,chen2022representative} utilize contrastive learning to improve the visual and textual representations by contrasting positive and negative pairs. They proposed various contrastive learning objectives to capture the abnormal regions from a chest X-Ray image. Since normal images usually dominate the dataset over abnormal ones~\cite{shin2016learning}, it is also crucial to recognize the normal or abnormal cases at the meantime.

In this paper, we propose a novel framework, named DCL, which exploits a dynamic graph integrating specific knowledge with general knowledge to enhance visual representations learned in an contrastive manner. We adopt the general knowledge with 28 entities from \cite{zhang2020radiology} as the fundamental structure of our graph, and the relationships are modelled in an adjacency matrix. Given a medical image, we first retrieve its semantically similar reports from the training set. Specific knowledge are extracted from those reports via RadGraph~\cite{jain2021radgraph} and stored in triplets (\textit{$<$subjective entity, relation, objective entity$>$}). And we integrate those triplets with the pre-constructed graph by dynamically adding additional nodes or linking two entities. 
We utilize a graph encoder to propagate information over the updated graph for refining the node features, which are initialized by a pretrained SciBert~\cite{beltagy2019scibert}.
Then the dedicated node features are attended to visual representations for report generation via a Transformer~\cite{vaswani2017attention} decoder. Based on the dynamic graph, we introduce a contrastive learning objective, image-report contrastive loss to well represent the visual features and textual information. In addition, contrastive learning can help ensure the accuracy for the report retrieval procedure in the dynamic graph construction process. Image-report matching loss is also employed to further improve the performances.

We evaluate our method on two benchmarks, IU-Xray~\cite{iuxray} and MIMIC-CXR~\cite{Johnson2019MIMICCXRAL}. Experimental results demonstrate that our approach can either outperform or match previous state-of-the-art (SOTA) methods in metrics that measure descriptive accuracy and clinical correctness. It indicates that leveraging dynamic graph to enhance contrastive learning is helpful to generate high-quality reports.

In summary, our main contributions are as follows:
\begin{itemize}
    \item We propose a novel framework that leverages a dynamic graph to enhance visual representations with contrastive learning paradigm for radiology reporting.
    \item Our proposed dynamic graph integrates both general and specific knowledge; The contrastive learning objective can improve visual and textual representations and dynamic graph accuracy.
    \item We conduct extensive experiments on two popular benchmarks to show the effectiveness of our approach, which achieves the SOTA performance on both language generation and clinical efficacy metrics.
\end{itemize}

\section{Related Work}
\label{sec:review}

\begin{figure*}[t]
  \centering
   \includegraphics[width=0.99\linewidth]{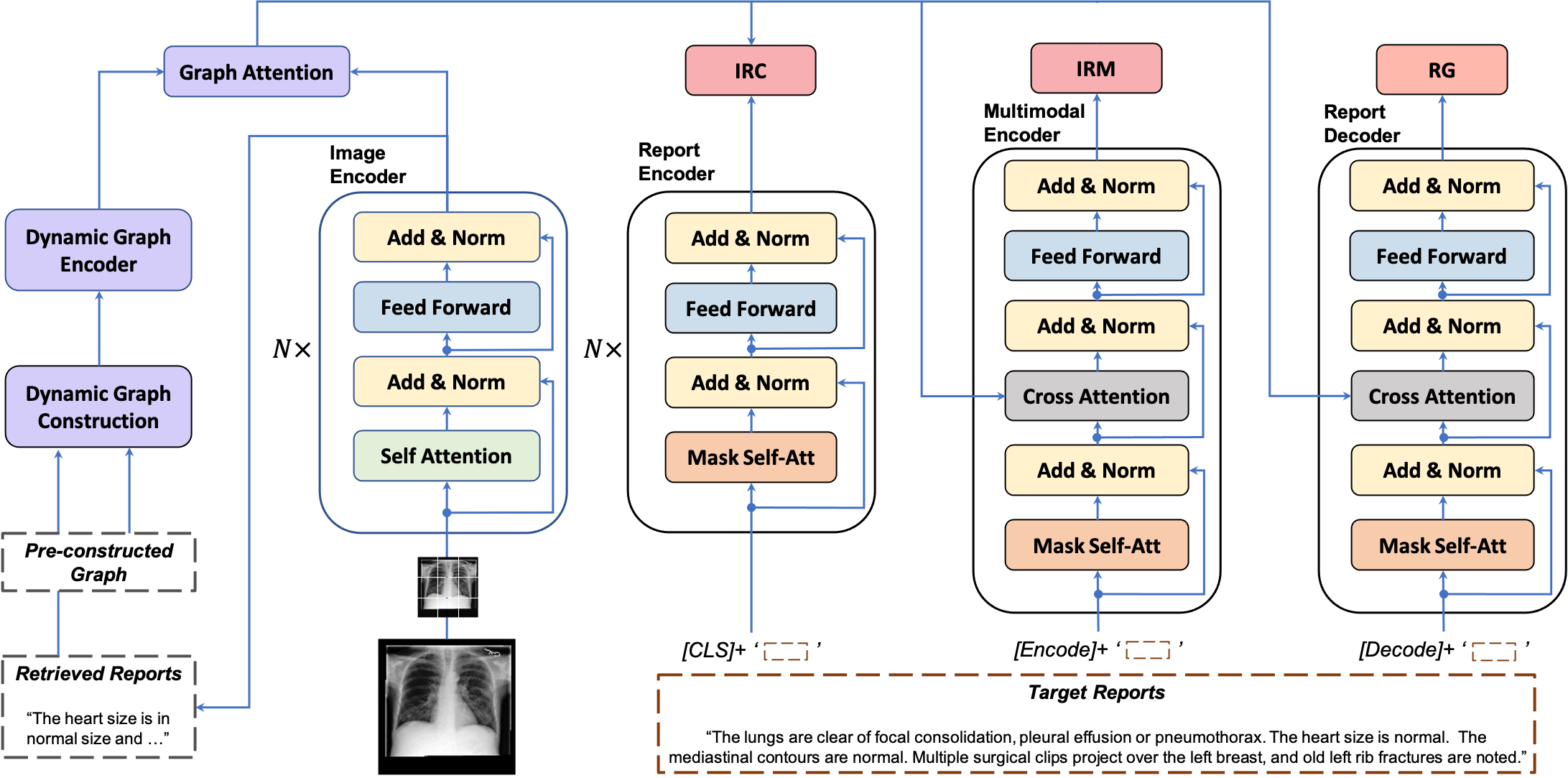}
   \caption{Illustration of our proposed \textbf{D}ynamic graph enhanced \textbf{C}ontrastive \textbf{L}earning approach (DCL). DCL contains two unimodal encoders, a multimodal encoder, a report decoder and three dynamic graph modules for construction, encoding and cross-modal attention, respectively. In addition to Report Generation (RG) loss, Image-Report Contrastive (IRC) loss and Image-Report Matching (IRM) loss are adopted for training DCL.}
   \label{fig:overview}
   \vspace{-0.4cm}
\end{figure*}

\subsection{Medical Report Generation Meets Knowledge Graph}
When radiologists write reports, they will make inferences with their expert knowledge. Typically, a report contains many sections, \eg, impressions, finding, comparison and indication. Following the previous works, we combine the impressions and finding sections as the target for report generation. To endow the MRG systems with capability to incorporate medical knowledge, various kinds of knowledge graphs have been explored and can be roughly divided into three groups. The first kind is proposed to emphasize the abnormal terminologies or disease keywords. Li \etal~\cite{li2019knowledge} collected abnormalities from MIMIC-CXR dataset, and utilized them as the node of their proposed abnormality graph. Edges here are the attention weights from source nodes to target nodes in the graph. Such graph is adopted in the following works~\cite{li2022auxiliary,zhao2021automatic} 
as prior medical knowledge to enhance the generation procedure and can even facilitate the unsupervised learning framework~\cite{liu2021auto}. Secondly, Zhang \etal~\cite{zhang2020radiology} and Liu \etal~\cite{liu2021exploring} adopted a universal graph with 20 entities. Entities linked to the same organ are connected to each other in the graph. Such relationships are modelled into a adjacency matrix and utilized to propagate messages in a graph encoding module. Since this graph is pre-constructed with prior knowledge in a fixed structure, we found that it can not guarantee the appropriate scope of knowledge for some cases (\eg. missing or unconnected common entities). To tackle this challenge, Liu \etal~\cite{liu2021exploring} utilized the global representations from pre-retrieved reports from training corpus to model domain-specific knowledge. 
In contrast, we aim to directly update the pre-constructed graph to model the appropriate knowledge. In the last, Li \etal~\cite{li2022cross} constructed a clinical graph by extracting structural information from training corpus via a NLP-rule based algorithm. They restored a set of triplets for each case to model the domain-specific knowledge and replace the visual representations. 

Our approach are based on the second category. Instead of using the fixed graph in \cite{zhang2020radiology}, we dynamically update the graph by injecting new knowledge extracted from the retrieved reports for each case. As a result, the appropriate scope of knowledge for different cases can be activated to generate more high-quality reports.

\subsection{Contrastive Learning}

The goal of contrastive learning is to improve representation learning by contrasting positive/negative or similar/dissimilar pairs. Inspired by the recent success of contrastive learning in vision-and-language pretraining tasks~\cite{li2022blip}, some works have introduced it in the MRG systems. Yan \etal~\cite{yan2021weakly} developed a weakly supervised method to contrast target reports with incorrect ones by identifying ``hard'' negative samples. To detect the abnormal regions, Liu \etal~\cite{liu2021contrastive} compared the referring image with known normal images via a contrastive attention mechanism. Other works~\cite{chen2022representative,zhang2020contrastive,wu2022multimodal} employed the contrastive learning during pretraining process to better represent visual features and textual information. All those works aimed to improve the expressiveness of both visual and textual representations and then facilitate radiology reporting. In our work, contrastive learning can also improve the accuracy of dynamic graph by training model to retrieve the most semantically similar reports.

\section{Methodology}
\label{sec:method}

In this section, we will introduce the detailed implementations of our proposed \textbf{D}ynamic graph enhanced \textbf{C}ontrastive \textbf{L}earning approach (DCL). The overall structure of DCL is illustrated in Fig.~\ref{fig:overview}, which contains four basic modules and three dynamic graph modules 
with
three training objectives. We first describe the background of DCL, and then introduce the dynamic graph modules and contrastive learning objectives, respectively. 

\subsection{Background}

\noindent\textbf{Notation} In this work, we aim to leverage dynamic graph to enhance contrastive learning for radiology reporting. In MRG task, the computer is asked to describe a given medical image $I$ 
with a free-text report $T = \{y_1,y_2,\dots,y_{n}\}$. 
We denote the target report by $\hat{T} = \{\hat{y}_1,\hat{y}_2,\dots,\hat{y}_{\hat{n}}\}$. $n$ and $\hat{n}$ represent the numbers of tokens in a report.
Our dynamic graph, 
denoted by
$G = \{V, E\}$, where $V$ and $E$ are the sets of nodes and edges, respectively, is built on a pre-constructed graph $G_{pre}$ proposed in \cite{zhang2020radiology} and updated with specific knowledge $K_{I}=\{k_{I}^{1},...,k_{I}^{n_{K}}\}$ extracted from retrieved reports $\{T^*_{i}\}_{i=1}^{n_{T}}$. 
Each $k$ is stored in a triplet format, which consists of a subjective entity $e_s$, an objective entity $e_o$ and their relation $r$. 
$n_{T}$ and $n_{K}$ are the numbers of reports and triplets, respectively.
All the triplets are acquired from the RadGraph~\cite{jain2021radgraph}. 

Typical MRG systems are encoder-decoder frameworks. The encoder is usually a CNN, \eg, ResNet~\cite{resnet} or DenseNet~\cite{huang2017densely}, encodes the given image $I$ to dense visual vectors $\mathbf{f}_I$. The decoder is usually a RNN (\eg, LSTM~\cite{hochreiter1997long}) or a Transformer~\cite{vaswani2017attention},
which decodes
$\mathbf{f}_I$ to a report $T$. In this work, we adopt the 
Transformer~\cite{vaswani2017attention} as the backbone to generate long and robust report.

\noindent\textbf{Image Encoder} 
Recently, visual Transformers have shown superior capability to represent images than CNNs. Thus, we only employ a ViT~\cite{dosovitskiy2020image} pretrained on the ImageNet~\cite{deng2009imagenet} as the image encoder to simplify the architecture. The input image will be divided into 196 patches, and a \texttt{[CLS]} token is further appended to the beginning of sequence before fed into the encoder layers. The whole process of an encoder layer $\mathbf{f}_{e}(\cdot)$ can be written as follows:
\begin{align}
    \mathbf{f}_e(x) & = \text{LN}(\text{FFN}(e_{attn})+e_{attn}), \\
    e_{attn} & = \text{LN}(\text{MHA}(x)+x),
\end{align}
where FFN and LN denote the Feed Forward Network~\cite{vaswani2017attention} and Layer Normalization operation~\cite{ba2016layer}, respectively. $x$ is the input of each encoder layer. MHA~\cite{vaswani2017attention} (multi-head attention) divides a scaled dot-product attention into $n$ parallel heads and each head $\text{Att}(\cdot)$ can be written as follows:
\begin{align}
    \text{Att}(x) = \text{softmax}(\frac{\mathbf{Q}^x(\mathbf{K}^x)^\top}{\sqrt{d}})\mathbf{V}^x,
\end{align}
where $d=768$ is the dimension of the embedding space and $\{ \mathbf{Q}, \mathbf{K^*}, \mathbf{V^*}\}$ are the packed $d$-dimensional \textit{Query, Key, Value} vectors, respectively. The final output is the encoded visual vectors $\mathbf{f}_I$, which will be used for report generation.

\noindent\textbf{Report Decoder} Our report decoder consists of two Transformer decoder layers. The whole process of a decoder layer $\mathbf{f}_{d}(\cdot)$ can be written as follows:
\begin{align}
    \mathbf{f}_d(\mathbf{y}) & = \text{LN}(\text{FFN}(e_{ca})+e_{ca}), \\
    e_{ca} & = \text{LN}(\text{CA}(e_{attn}, \mathbf{f}_{I})+e_{attn}), \\
    e_{attn} & = \text{LN}(\text{MMHA}(\mathbf{y})+\mathbf{y})),
\end{align}
where MMHA and CA represents the 
masked multi-head self-attention and cross attention mechanism in \cite{vaswani2017attention}. $\mathbf{y}$ is the input of decoder. A \texttt{[Decode]} token is added to the beginning of $\mathbf{y}$ to signal the start while a \texttt{[EOS]} token is to signal its end. 
In Cross-attention sublayer, for each head, $\{ \mathbf{Q}, \mathbf{K^*}, \mathbf{V^*}\}$ comes from $\mathbf{Q} = W_{q}*e_{attn}$, $\mathbf{K} = W_{k}*\mathbf{f}_I$, and $\mathbf{V} = W_{v}*\mathbf{f}_I$, where $W_{*}$ are the learnable parameters. 
The $f_d(\mathbf{y})$ will be sent to a Linear $\&$ Log-Softmax layer to get the output of target sentences. Notably, only token embedding is adopted during the decoding procedure. The entire 
auto-regressive generation process can be written as follows:
\begin{align}
    p(T|I) = \prod_{t=1}p(y_t|y_1,\dots,y_{t-1}, I).
\end{align}
where $y_t$ is the input token in time step $t$.

Typically, the report generation objective is the cross-entropy loss to compare the predicted token index sequence with the ground truth. Given the ground truth report $\hat{T}$, all the underlying modules are trained to maximize $p(\mathbf{y}|I)$ by minimizing the following:
\begin{align}
    \mathcal{L}_{\mathrm{RG}}= - \sum_{t=1}^{\hat{n}}\log p(\hat{y}_t|\hat{y}_1,\cdots,\hat{y}_{t-1}, I).
\end{align}

\subsection{Dynamic Graph}
\label{sec:dynamic grah}

\begin{figure}[t]
  \centering
   \includegraphics[width=0.9\linewidth]{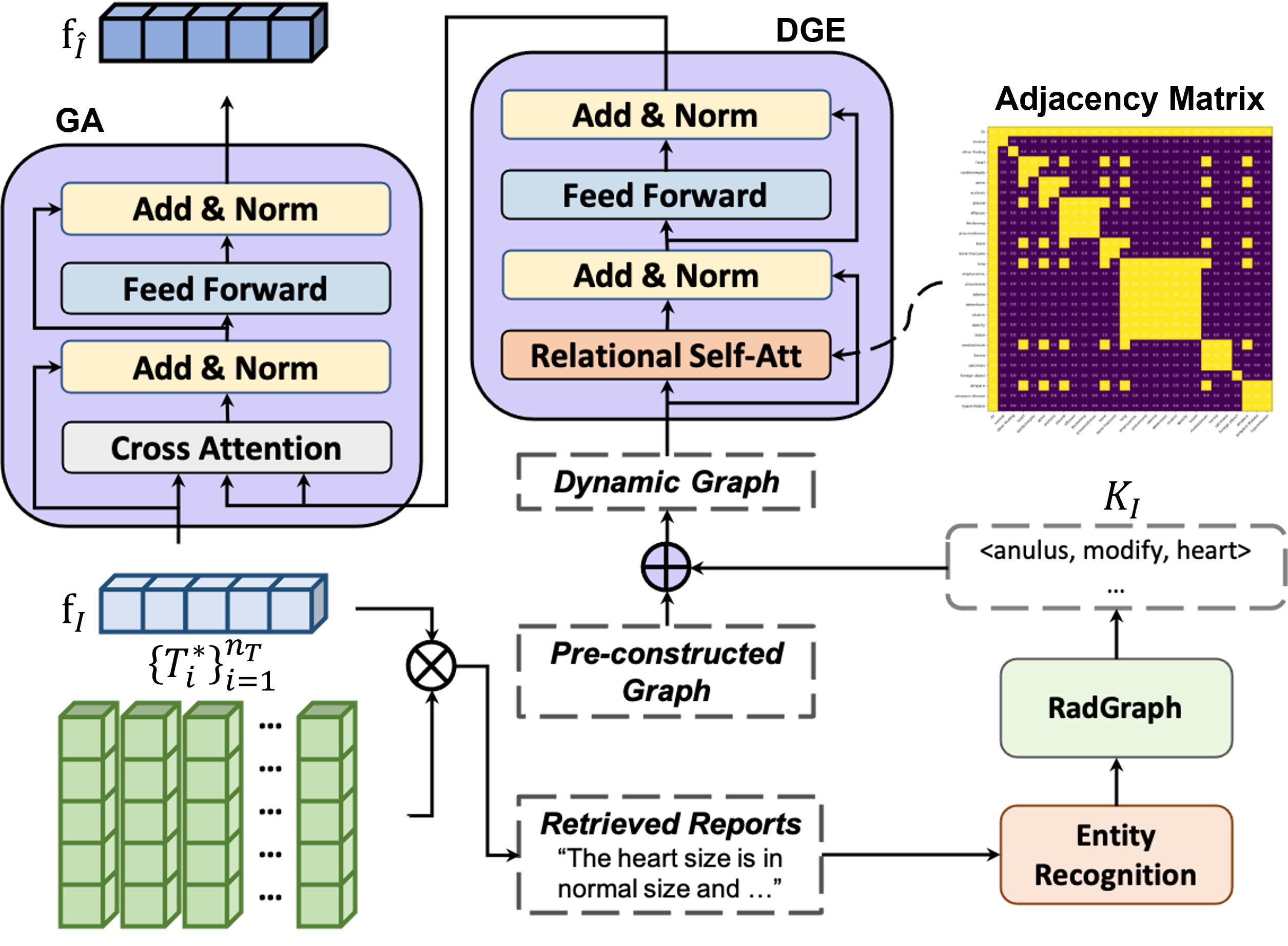}
   \caption{illustration of our proposed dynamic graph construction, dynamic graph encoder (DGE) and graph attention (GA) modules. The structure of pre-constructed graph can be found in \cref{fig:motivation}. 
   }
   \label{fig:dynamic}
   \vspace{-0.4cm}
\end{figure}

The chest knowledge graph 
$G_{pre}$ proposed in~\cite{zhang2020radiology} 
has been widely integrated with MRG systems to emphasize the disease keywords and enhance their relationships. $G_{pre}$ consists of 27 entities and a root node referring to the global feature and an adjacency matrix $A = \{e_{ij}\}$ to represent the edges $V$. Each node is a disease keyword and we set $e_{ij}$ to 1, when source node $n_i$ connects target node $n_j$. Nodes linked to the same organ or tissue are connected to each other and the root. This graph is not updated during the training, and we found that it limits the effectiveness from two aspects. 
Firstly, those entities can not cover most common disease keywords for all datasets because of the dataset bias;
Secondly, entities linked to different organs can also affect each other clinically. To tackle those limitations, we propose a dynamic graph $G$ with dynamic structure and nodes and integrate it with visual features to generate high-quality reports. This process is illustrated in Fig.~\ref{fig:dynamic}, and we will introduce the three key modules, i.e., dynamic graph construction, dynamic graph encoder, and graph attention in this section, respectively.

\noindent\textbf{Dynamic Graph Construction} We construct our graph in a bottom-up manner, in which we first construct the fundamental structure from general knowledge and then add nodes or redefine their relationships according to specific knowledge. We extend $G_{pre}$ to our fundamental structure with 28  entities consisting of a global node represented by a \texttt{[CLS]} token, 7 organs or tissues, and 20 disease keywords. 
In addition to link criteria in $G_{pre}$, every organ will connect to each other organ or tissue.

To get the specific knowledge for each given image $I$, we first retrieve the 
top-$n_{T}$
similar reports 
$\{T^*_{i}\}_{i=1}^{n_{T}}$
by calculating the similarity between the visual feature $\mathbf{f}_I$ and representations of reports in queue $\{\mathbf{f}_{T^*}^i\}_{i=1}^{n_Q}$, where $n_Q$ is the length of 
the report queue. For each retrieved report, we extract anatomy and observation entities by Stanza~\cite{zhang2021biomedical}. Those entities are further utilized to quote the specific knowledge $K_{I}$ from RadGraph~\cite{jain2021radgraph}. Each triplet $k$ in $K_{I}$ aims to depict the relationship between source and target entities. There are three kinds of relations in RadGraph, namely \textit{`suggestive of'}, \textit{`modify'} and \textit{`located at'}. 
For a triplet whose only source entity $e_s$ or target entity $e_o$ is in the graph, we will add another entity in the triplet as an additional node, and set their relation to 1 in the adjacency matrix $A$. Note that  if the relation is \textit{`located at'} and the entity needed to be add is the target entity $e_o$, $e_o$ will be treated as an organ/tissue node.
In this bottom-up manner, our dynamic graph is capable to exploit both general and specific knowledge.

\noindent\textbf{Dynamic Graph Encoder} We propose a dynamic graph encoder to propagate information and learn dedicated node features in our dynamic graph. To this end, this module is built upon the standard Transformer encoder layer $\mathbf{f}_G$ and conducted as (see \Cref{fig:dynamic}):
\begin{align}
    \mathbf{f}_G & = \text{LN}(\text{FFN}(e_{rsa})+e_{rsa}), \\
    e_{rsa} & = \text{LN}(\text{RSA}(\mathbf{f}_N,A)+\mathbf{f}_N),
\end{align}
where RSA is a MMHA-like relational self attention module to encode the structural information of a graph to the model. Concretely, we utilize adjacency matrix $A$ as a visible mask~\cite{liu2020k,li2022cross} to control the original MMHA.
It promises that each node can only impact its linked nodes and enhance their relationships. $\mathbf{f}_N$ is the initialized nodes representations and consists of entity embedding and level encoding. Previous works initialized representations randomly, which limits the effectiveness seriously~\cite{zheng2022rethinking}. Moreover, some additional nodes appear few times during the training and hard to find the best embeddings. Therefore, we first adopt word embeddings $\epsilon_{sci}$ from well-pretrained SciBert~\cite{beltagy2019scibert} to initialize each entity. For those entities consist of one more words, \eg, `foreign object', we calculate the average. Furthermore, we add level encoding $\epsilon_{l}$ to demonstrate each node is the root, organ or a disease keyword. Thus, the structural information of our graph is well represented and encoded during message passing and propagation.

\noindent\textbf{Graph Attention} Graph attention aims to integrate knowledge from dynamic graph with visual features. Following \cite{liu2021exploring}, we utilize cross attention to achieve this goal. The whole process can be written as follows:
\begin{align}
    \mathbf{f}_{\hat{I}} & = \text{LN}(\text{FFN}(e_{ga})+e_{ga}), \\
    e_{ga} & = \text{LN}(\text{CA}(\mathbf{f}_{I},  \mathbf{f}_G))+\mathbf{f}_{I}).
\end{align}
In each head, \textit{Query} comes from visual features $\mathbf{f}_{I}$ while \textit{Key} and \textit{Value} come from the learned graph representations $\mathbf{f}_G$. Finally, we get the dynamic graph enhanced visual features $\mathbf{f}_{\hat{I}}$. Notably, the first token in both $\mathbf{f}_{\hat{I}}$ and $\mathbf{f}_G$ is \texttt{[CLS]} to aggregate visual and graph information.

\subsection{Contrastive Learning}
Collecting paired image and text data is prohibitively expensive leading to smaller size of MRG datasets compared with captioning datasets, like COCO~\cite{lin2014microsoft}. It hinders the potential of existing data-driven MRG systems. 
In this section, we introduce the image-report contrastive loss used in our DCL, which can effectively improve the visual and textual representations as well as ensure the report retrieval accuracy in the dynamic graph construction process. Moreover, inspired by recent vision-language pretraining works~\cite{li2022blip,li2021align}, we also adopt an image-report matching loss in the proposed method for further enhancing the representations to improve the performance.

\noindent\textbf{Image-Report Contrastive Loss} (IRC) can activate radiology reporting by encouraging the positive image-report pairs to have similar representations in contrast to the negative pairs. A report encoder with the same architecture with image encoder is utilized to extract textual representations $\mathbf{f}_{T}$ referring to the positive or negative report. Then we calculate the similar between two \texttt{[CLS]} representations by $s = W_I(V_{cls})^tW_T(T_{cls})$, where $W_I$ and $W_T$ are two learnable matrices. we also maintain two queues to store the most recent $M$ image-report representations from the momentum image and report encoders. After a Softmax activation, we can get the image-report similarity $f_m^{i2r}(I) = \frac{\exp s(I,T_m)\//\tau}{\sum_{m=1}^M\exp s(I,T_m)\//\tau}$ and the report-image similarity $f_m^{r2i}(T)$, where $\tau$ is a learnable temperature parameter. The IRC can be written as follows:
\begin{align}
    \mathcal{L}_{\mathrm{IRC}}= \frac{1}{2}(\mathcal{L}_{\mathrm{ce}}(g(T),f(T))+\mathcal{L}_{\mathrm{ce}}(g(I),f(I))),
\end{align}
where $\mathcal{L}_{\mathrm{ce}}$ is the cross entropy loss and $g(I)$ is the ground truth of image-report similarity.

\noindent\textbf{Image-Report Matching Loss} (IRM) is a binary classification task to predict whether the given image-report pair is positive (matched) or negative (unmatched). Different from IRC, we utilize a multimodal encoder to capture the multimodal representations via cross attention mechanism. Then \texttt{[Encode]} vector is projected to $d=2$ with a linear layer to predict the probability $p^{itm}$. The IRM is conducted as:
\begin{align}
    \mathcal{L}_{\mathrm{IRM}} = \mathcal{L}_{\mathrm{ce}}(g^{itm}, p^{itm}).
\end{align}
Finally, we calculate the sum of $\mathcal{L}_{\mathrm{RG}}$, $\mathcal{L}_{\mathrm{IRC}}$ and $\mathcal{L}_{\mathrm{IRM}}$ as our total loss function.
Notably, the multimodal encoder is only used during the training process to improve the representation learning.

\vspace{-0.2cm}
\section{Experiments}
\vspace{-0.2cm}

\subsection{Datasets, Evaluation Metrics and Settings}

We evaluate our proposed DCL on two widely-used radiology reporting benchmarks, IU-Xray~\cite{iuxray} and MIMIC-CXR~\cite{Johnson2019MIMICCXRAL}. We adopt the settings in \cite{liu2021exploring,R2gen} to split those two datasets and preprocess the reports for a fair comparison.

\noindent\textbf{IU-Xray} \cite{iuxray} has been widely used to evaluate the performance of radiology reporting systems. It contains 3,955 radiology reports and 7,470 chest xray images. Either frontal or frontal and lateral view images are associated with each report. Following \cite{R2gen,li2019knowledge}, we exclude those cases with only one image and finally get 2069/296/590 cases for training/validation/testing. By Stanza~\cite{zhang2021biomedical}, 739 unique entities are extracted and utilized as dynamic nodes candidates.

\noindent\textbf{MIMIC-CXR} \cite{Johnson2019MIMICCXRAL} is the largest radiology dataset to date, consists of 368,960 chest X-ray images and 222,758 radiology reports and is splitted officially. Recently, various MIMIC-child datasets have been proposed by exploring structural radiology information, \eg, Chest ImaGenome~\cite{wuchest} and RadGraph~\cite{jain2021radgraph}. In this work, we adopt RadGraph to update our dynamic structure. Classified by relation, RadGraph consists of 2,895,725 (\textit{suggestive of}), 6,115,264 (\textit{located at}) and 4,010,875 (\textit{modify}) triplets.

\noindent\textbf{Natural Language Generation Metrics} (NLG) are used to measure descriptive accuracy of predicted reports. CIDEr~\cite{vedantam2015cider} and BLEU~\cite{papineni-etal-2002-bleu} are two main NLG metrics used to measure the quality of predicted reports. BLEU is proposed for machine translation tasks and measures the word n-gram overlap between predictions and reports. Due to the textual bias in MRG datasets, MRG systems can achieve considerable BLEU values even when they just repeat the most frequent sentences. In contrast, CIDEr is tailored to evaluate captioning systems by rewarding topic terms (terminologies in MRG task) and penalizing frequent terms. Additionally, values of ROUGE-L~\cite{lin-2004-rouge} and METEOR~\cite{banerjee2005meteor} are also reported for comparison. 

\noindent\textbf{Clinical Efficacy Metrics} are recently proposed to capture and evaluate clinical correctness of predicted reports. It first employs the CheXPert labeling tool proposed in \cite{irvin2019chexpert} to label predicted reports and the ground truth reports in 14 different medical terminologies. Then classification measurements, i.e., F1-Score, Precision and Recall are calculated to evaluate how well the generated report describes the abnormalities. Since the provider of IU-Xray does not use CheXPert to build the labels, CE metrics are only reported on the MIMIC-CXR dataset~\cite{yang2022knowledge,song2022cross}. 

\noindent\textbf{Experimental Settings} We use the same image encoder for different views images, and concatenate visual tokens via fusion operation for further process. Considering the domain gap between medical and generic texts, we employ a pretrained SciBert~\cite{beltagy2019scibert} to serve as tokenizer and report encoder. The model is trained on 4 NVIDIA 2080 Ti GPUs with the batch size 8 and 30 epochs. The checkpoint acquires the highest CIEDr metric is used for testing. The learning rate is set as 1e-4 and the optimizer is AdamW~\cite{loshchilov2017decoupled} with a weight decay of 0.02. Top 3 similar reports from the text queue $Q$ are retrieved. And the size of $Q$ is set as 65,536 and 1,380 for MIMIC-CXR and IU-Xray. The max length of specific knowledge is set as 90. For batch operation, the nodes in $G$ is padded to 50 with a \texttt{[PAD]} token. Note that, we project all encoded vectors by a linear transformation layer into the dimension of $d=768$.

\begin{table*}
\centering
\resizebox{\textwidth}{!}{
\begin{tabular}{lcccc|lcccc} 
\toprule
\multicolumn{5}{c}{IU-Xray\cite{iuxray}}                                                             & \multicolumn{5}{c}{MIMIC-CXR\cite{Johnson2019MIMICCXRAL}}                                                              \\ 
\midrule
\midrule
Methods & CIDEr                 & BLEU-4         & ROUGE-L        & METEOR                 & Methods   & CIDEr                 & BLEU-4         & ROUGE-L        & METEOR                 \\ 
\midrule
R2Gen\cite{R2gen} & 0.398  & 0.165          & 0.371          & 0.187                  & R2Gen\cite{R2gen}   & 0.253 & 0.103          & 0.277          & 0.142                  \\
KERP\cite{li2019knowledge}  & 0.280                 & 0.162          & 0.339          & \multicolumn{1}{c|}{-} & CMN\cite{chen2022cross}     & \multicolumn{1}{c}{-} & 0.106          & 0.278          & 0.142                  \\
HRGP\cite{Li2018HybridRR}  & 0.343                 & 0.151          & 0.322          & \multicolumn{1}{c|}{-} & TopDown\cite{anderson2018bottom} & 0.073                 & 0.092          & 0.267          & 0.129                  \\
MKG\cite{zhang2020radiology}   & 0.304                 & 0.147          & 0.367          & \multicolumn{1}{c|}{-} & M2TR\cite{nooralahzadeh2021progressive}    & \multicolumn{1}{c}{-} & 0.107          & 0.272          & 0.145                  \\
PPKED\cite{liu2021exploring} & 0.351                 & 0.168          & 0.376          & 0.190                  & PPKED\cite{liu2021exploring}   & 0.237                 & 0.106          & \textbf{0.284} & 0.149                  \\
MGSK\cite{yang2022knowledge}  & 0.382                 & \textbf{0.178} & 0.381          & \multicolumn{1}{c|}{-} & MGSK\cite{yang2022knowledge}    & 0.203                 & \textbf{0.115} & \textbf{0.284} & \multicolumn{1}{c}{-}  \\
CA\cite{liu2021contrastive}    & \multicolumn{1}{c}{-} & 0.169          & 0.381          & \textbf{0.193}         & CA\cite{liu2021contrastive}      & \multicolumn{1}{c}{-} & 0.109          & 0.283          & \textbf{0.151}         \\
CMCL\cite{liu2022competence}  & \multicolumn{1}{c}{-} & 0.162          & 0.378          & 0.186                  & CMCL\cite{liu2022competence}    & \multicolumn{1}{c}{-} & 0.097          & 0.281          & 0.133                  \\ 
\midrule
Ours  & \textbf{0.586}        & 0.163          & \textbf{0.383} & \textbf{0.193}         & Ours    & \textbf{0.281}        & 0.109          & \textbf{0.284} & 0.150                  \\
\bottomrule
\end{tabular}}
\caption{The performances of our proposed DCL compared with other state-of-the-art systems on IU-Xray and MIMIC-CXR dataset. The best results in each column are highlighted in bold. CIDEr~\cite{vedantam2015cider} is proposed to evaluate captioning systems.}
\label{table:main results}
\vspace{-0.4cm}
\end{table*}

\begin{table}
\centering
\begin{tabular}{l|ccc} 
\toprule
Methods      & Precision & Recall & F1-score  \\ 
\midrule
TopDown~\cite{anderson2018bottom}      & 0.166     & 0.121  & 0.133     \\
M2TR~\cite{nooralahzadeh2021progressive}         & 0.240     & 0.428  & 0.308     \\
R2Gen~\cite{R2gen}        & 0.333     & 0.273  & 0.276     \\
MKSG~\cite{yang2022knowledge}         & 0.458     & 0.348  & 0.371     \\ 
\midrule
Ours w/o $G$    & 0.275     & 0.185  & 0.194     \\
Ours w/o \small$\mathbf{L}_{IRC}$ & 0.463     & 0.337  & 0.359    \\
Ours w/o \small$\mathbf{L}_{IRM}$ & 0.469     & \textbf{0.353}  & 0.372      \\
Ours         & \textbf{0.471}     & 0.352  & \textbf{0.373}     \\
\bottomrule
\end{tabular}
\caption{The comparison of the clinical efficacy metrics on MIMIC-CXR dataset. The w/o is the abbreviation of without.}
\label{table:ce}
\vspace{-0.4cm}
\end{table}

\subsection{Main Results}

\noindent\textbf{Descriptive Accuracy} In \cref{table:main results}, we compare our DCL with a wide range of existing state-of-the-art MRG systems on two benchmarks. R2Gen~\cite{R2gen} and CMN~\cite{chen2022cross} have been widely used a baseline MRG model recently. KERP~\cite{li2019knowledge}, MKG~\cite{zhang2020radiology}, PPKED~\cite{liu2021exploring} and MGSK~\cite{yang2022knowledge} are proposed to integrate medical knowledge with typical MRG backbones. CA~\cite{liu2021contrastive} and CMCL~\cite{liu2022competence} employ contrastive learning and curriculum learning to improve the performance. The performances of other baseline models, such as HRGP~\cite{Li2018HybridRR}, M2TR~\cite{nooralahzadeh2021progressive} and TopDown~\cite{anderson2018bottom} are also reported. Since we follow the same settings, we directly cite the results from original papers. As shown in \cref{table:main results}, our DCL achieves the SOTA descriptive accuracy, which outperforms others in CIDEr and ROUGE-L metrics and match their performances in BLEU-4 and METEOR metrics. Higher CIDEr values demonstrate that our model does not repeat frequent sentences in training set but generating reports with more accurate topics.

\begin{figure}[t]
  \centering
   \includegraphics[width=0.9\linewidth]{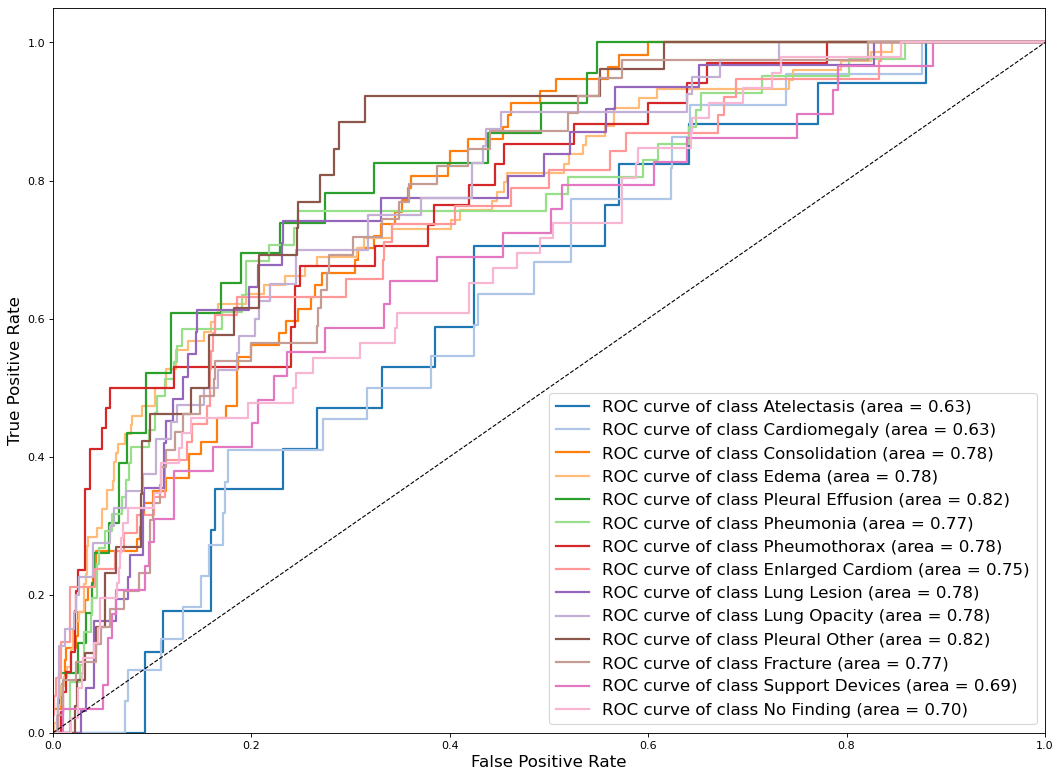}
   \caption{Micro-average of receiver operating characteristic curve for clinical abnormalities predictions from the generated reports.}
   \label{fig:roc}
   \vspace{-0.4cm}
\end{figure}

\noindent\textbf{Clinical Correctness} We also evaluate our method by clinical efficacy (CE) metrics on the MIMIC-CXR dataset and compare the performances with other baseline models. Following official splitting, we directly cite the results from \cite{yang2022knowledge} for comparison. In \cref{fig:roc}, we show the micro-average of ROC for 14 chesty terminologies prediction and present the AUC scores. `Pleural Effusion' and `Pleural Other' achieve the highest AUCs (0.82). The experimental results in \cref{table:ce} reveal that our DCL significantly outperforms the previous models on three CE metrics. Compared with the current SOTA method MGSK~\cite{yang2022knowledge} that also leverage general prior knowledge and specific knowledge from RadGraph~\cite{jain2021radgraph}, we make a performance-boosting. The improvement verifies the importance of our dynamic graph concepts and also demonstrates that our system can predict more accurate clinical information. 

\subsection{Ablation Study}

In this section, we conduct ablation studies on IU-Xray and MIMIC-CXR datasets to investigate the contribution of each component in our proposed DCL. 
Tab.~\ref{tab:ablationstudy} presents the quantitative analysis of DCL on IU-Xray with measuring descriptive accuracy. And clinical correctness evaluation is reported in \cref{table:ce}. Our base model only keeps the image encoder and report decoder and employs $\mathcal{L}_{\mathrm{RG}}$.

\begin{table*}
\centering
\begin{tabular}{l|cccccc|cccc} 
\toprule
Settings & $G_{pre}$  & $K_{I}$  & IRC & IRM & ViT\cite{dosovitskiy2020image} & SciBert\cite{beltagy2019scibert} & CIDEr & BLEU-4 & ROUGE-L & METEOR  \\ 
\midrule
\midrule
Base     &     &     &     &     & \checkmark & \checkmark     & 0.383 & 0.133  & 0.277   & 0.163   \\ 
\midrule
(a)      & \checkmark &     &     &     & \checkmark & \checkmark     & 0.535 & 0.144  & 0.349   & 0.180   \\
(b)      & \checkmark & \checkmark &     &     & \checkmark & \checkmark     & 0.557 & 0.150  & 0.361   & 0.182   \\ 
\midrule
(c)      & \checkmark & \checkmark & \checkmark &     & \checkmark & \checkmark     & 0.580 & 0.161  & \textbf{0.385}   & 0.188   \\
(d)      & \checkmark & \checkmark &     & \checkmark & \checkmark & \checkmark     & 0.564 & 0.155  & 0.370   & 0.188   \\ 
\midrule
(e)      & \checkmark & \checkmark & \checkmark & \checkmark & \checkmark &         &   0.580   &   0.158  &   0.374     &   0.190     \\
(f)      & \checkmark & \checkmark & \checkmark & \checkmark &     & \checkmark     &    0.527    &     0.158      &     0.356     &    0.179      \\ 
\midrule
DCL      & \checkmark & \checkmark & \checkmark & \checkmark & \checkmark & \checkmark     & \textbf{0.586} & \textbf{0.163}  & 0.383   &\textbf{0.193}   \\
\bottomrule
\end{tabular}
\caption{Quantitative analysis of proposed method on IU-Xray dataset. The base model consists of an image encoder and a report decoder with report generation loss only.}
\label{tab:ablationstudy}
\vspace{-0.4cm}
\end{table*}

\begin{figure*}[t]
  \centering
   \includegraphics[width=\linewidth]{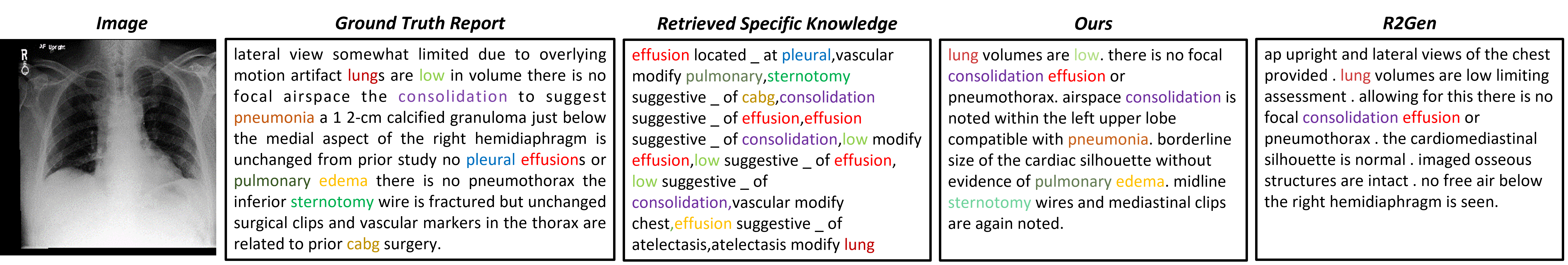}
   \caption{Illustrations of reports from ground truth, ours and R2Gen~\cite{R2gen} and retrieved specific knowledge for one sample from MIMIC-CXR~\cite{Johnson2019MIMICCXRAL}. For better visualization, different colors highlight different medical entities.}
   \label{fig:case study}
   \vspace{-0.4cm}
\end{figure*}

\noindent\textbf{Effect of Dynamic Graph} Our dynamic graph is constructed in a bottom-up manner, that exploits general knowledge from $G_{pre}$ and specific knowledge $K_{I}$ extracted from retrieved Top-3 similar reports. Comparing the base model with setting (a) and (b), our dynamic graph can boost the performance of base model, substantially. More specifically, leveraging the general knowledge only lead to an increase on all NLG metrics by 15.2$\%$ on CIDEr and 1.1$\%$ on BLEU-4. By integrating specific knowledge, our dynamic graph can further boost the performances, \eg~0.535 $\rightarrow$ 0.557 on CIDEr. It demonstrates the effectiveness and necessity of constructing dynamic graph for each image. We hypothesize that this performance improvement may be due to that the dynamic knowledge can emphasise keywords when generating reports, since dynamic nodes are from retrieved reports. It has been proved in \cref{table:ce}, leveraging dynamic graph significantly improve the performances on all CE metrics, which means the generated reports can provide more accurate medical terminologies.

\noindent\textbf{Effect of Contrastive Learning} Sequentially, we evaluate the effectiveness of two introduced learning objectives, i.e., image report contrastive loss (IRC) and image report matching loss (IRM). The performances of settings (b,c) in \cref{table:ce} shows that both IRC and IRM can boost the base model performances. It proves the importance of visual and textual representation qualities, since serve data bias in MRG datasets will degenerate the representation capabilities seriously. However, comparing (c) and (d), it is observed that IRC brings more improvement than IRM. We speculate the reason is that IRC can straightly improve visual and textual representations by aligning similar pairs. In contrast, IRM works on multimodal representations and represents unimodal in an indirect manner.

\noindent\textbf{Choice of Parameter Initialization} We employ a pretrained ViT~\cite{dosovitskiy2020image} and SciBert~\cite{beltagy2019scibert} as image and report encoder in our implementation. It is worth noting that Transformers lack some of the inductive biases inherent to CNNs and therefore generalize bad with insufficient and unbalanced data~\cite{dosovitskiy2020image}. Not surprisingly, comparing setting (f) and full model, the performances drop steeply without pretrained ViT parameters. The performances of setting (e) demonstrates the effectiveness of pretrained SciBert. The improvement comes from two aspects. Firstly, it provides well pretrained parameters; Secondly, its tokenizer encodes each token (medical terminology) with a well pretrained embedding, which avoid Graph Transformers to propagate similar hidden states~\cite{zheng2022rethinking}.

\vspace{-0.1cm}
\subsection{Case Study}
\vspace{-0.1cm}
To further investigate the effectiveness of our method, we perform qualitative analysis on MIMIC-CXR~\cite{Johnson2019MIMICCXRAL} with their retrieved specific knowledge, and reports from ground truth, our model and R2Gen~\cite{R2gen}. Entities extracted by Stanza~\cite{zhang2021biomedical} from the ground truth report have been highlighted with different colors. It is observed that some entities, \eg~\textit{cabg}, \textit{consolidation}, and \textit{sternotomy}) are not included in the pre-constructed graph node lists. This observation proves our motivation for constructing knowledge graph dynamically. We conduct the same operation on retrieved reports and use extracted entities to quote related triplets from RadGraph~\cite{jain2021radgraph}. Those triplets are known as specific knowledge in this paper and shown in \cref{fig:case study}. Retrieved specific knowledge triplets \textit{$<$sternotomy,suggestive$\_$of,cabg$>$} and \textit{$<$consolidation,suggestive$\_$of,effusion$>$} demonstrate that the retrieved reports contain the similar medical terminologies and clinical information. We speculate that IRC and IRM objectives bring such capabilities. Then the entities in our dynamic graph emphasise disease/organ keywords when generating reports and it is why our DCL can predict sentence \textit{``airspace consolidation is noted within the left upper lobe compatible with pneumonia."}, but R2Gen can not. 

\vspace{-0.1cm}
\section{Conclusion and Discussion}
\vspace{-0.1cm}
In this paper, we present a practical approach to leverage dynamic graph to enhance contrastive learning for radiology report generation. In which the dynamic graph is constructed in a bottom-up manner to integrate retrieved specific knowledge with general knowledge. Then contrastive learning is employed to improve visual and textual representations, which also promises the accuracy of our dynamic graph. Experiments on two popular benchmarks verify the effectiveness of our method in generating accurate and meaningful reports. More encouragingly, our approach can outperform or match existing SOTA methods in language generation and clinical efficacy metrics.

\noindent\textbf{Limitation and Future Work} Retrieved reports can not be exactly the same as ground truth, and knowledge noises are involved during the dynamic graph construction process. It may guide the model to generate inaccurate sentences. In future, we plan to propose a specific objective for dynamic graph construction process to further improve the accuracy of dynamic graph and the quality of predicted reports.

\vspace{-0.1cm}
\section*{Acknowledgements}
\vspace{-0.1cm}

This work was partially supported by ``Taishan Scholars Youth Expert Program'' of Shandong Province, and partially supported by National Key R$\&$D Program of China under Grant No. 2020AAA0109700.


\clearpage
{\small
\bibliographystyle{ieee_fullname}
\bibliography{main}

\begin{thebibliography}{10}\itemsep=-1pt

\bibitem{anderson2018bottom}
Peter Anderson, Xiaodong He, Chris Buehler, Damien Teney, Mark Johnson, Stephen
  Gould, and Lei Zhang.
\newblock Bottom-up and top-down attention for image captioning and visual
  question answering.
\newblock In {\em Proceedings of the IEEE Conference on Computer Vision and
  Pattern Recognition}, pages 6077--6086, 2018.

\bibitem{ba2016layer}
Jimmy~Lei Ba, Jamie~Ryan Kiros, and Geoffrey~E Hinton.
\newblock Layer normalization.
\newblock {\em arXiv preprint arXiv:1607.06450}, 2016.

\bibitem{banerjee2005meteor}
Satanjeev Banerjee and Alon Lavie.
\newblock Meteor: An automatic metric for mt evaluation with improved
  correlation with human judgments.
\newblock In {\em Proceedings of the acl workshop on intrinsic and extrinsic
  evaluation measures for machine translation and/or summarization}, pages
  65--72, 2005.

\bibitem{beltagy2019scibert}
Iz Beltagy, Kyle Lo, and Arman Cohan.
\newblock Scibert: A pretrained language model for scientific text.
\newblock In {\em Proceedings of the Conference on Empirical Methods in Natural
  Language Processing}, pages 3615--3620, 2019.

\bibitem{chen2022representative}
Yu-Jen Chen, Wei-Hsiang Shen, Hao-Wei Chung, Jing-Hao Chiu, Da-Cheng Juan,
  Tsung-Ying Ho, Chi-Tung Cheng, Meng-Lin Li, and Tsung-Yi Ho.
\newblock Representative image feature extraction via contrastive learning
  pretraining for chest x-ray report generation.
\newblock {\em arXiv preprint arXiv:2209.01604}, 2022.

\bibitem{chen2022cross}
Zhihong Chen, Yaling Shen, Yan Song, and Xiang Wan.
\newblock Cross-modal memory networks for radiology report generation.
\newblock {\em arXiv preprint arXiv:2204.13258}, 2022.

\bibitem{R2gen}
Zhihong Chen, Yan Song, Tsung{-}Hui Chang, and Xiang Wan.
\newblock Generating radiology reports via memory-driven transformer.
\newblock In {\em Proceedings of the Conference on Empirical Methods in Natural
  Language Processing}, 2020.

\bibitem{iuxray}
Dina Demner-Fushman, Marc~D Kohli, Marc~B Rosenman, Sonya~E Shooshan, Laritza
  Rodriguez, Sameer Antani, George~R Thoma, and Clement~J McDonald.
\newblock Preparing a collection of radiology examinations for distribution and
  retrieval.
\newblock {\em Journal of the American Medical Informatics Association},
  23(2):304--310, 2016.

\bibitem{deng2009imagenet}
Jia Deng, Wei Dong, Richard Socher, Li-Jia Li, Kai Li, and Li Fei-Fei.
\newblock Imagenet: A large-scale hierarchical image database.
\newblock In {\em Proceedings of the IEEE Conference on Computer Vision and
  Pattern Recognition}, pages 248--255, 2009.

\bibitem{dosovitskiy2020image}
Alexey Dosovitskiy, Lucas Beyer, Alexander Kolesnikov, Dirk Weissenborn,
  Xiaohua Zhai, Thomas Unterthiner, Mostafa Dehghani, Matthias Minderer, Georg
  Heigold, Sylvain Gelly, et~al.
\newblock An image is worth 16x16 words: Transformers for image recognition at
  scale.
\newblock {\em arXiv preprint arXiv:2010.11929}, 2020.

\bibitem{endo2021retrieval}
Mark Endo, Rayan Krishnan, Viswesh Krishna, Andrew~Y Ng, and Pranav Rajpurkar.
\newblock Retrieval-based chest x-ray report generation using a pre-trained
  contrastive language-image model.
\newblock In {\em Machine Learning for Health}, pages 209--219, 2021.

\bibitem{resnet}
Kaiming He, Xiangyu Zhang, Shaoqing Ren, and Jian Sun.
\newblock Deep residual learning for image recognition.
\newblock In {\em Proceedings of the IEEE Conference on Computer Vision and
  Pattern Recognition}, pages 770--778, 2016.

\bibitem{hochreiter1997long}
Sepp Hochreiter and J{\"u}rgen Schmidhuber.
\newblock Long short-term memory.
\newblock {\em Neural computation}, 9(8):1735--1780, 1997.

\bibitem{hossain2019comprehensive}
MD~Zakir Hossain, Ferdous Sohel, Mohd~Fairuz Shiratuddin, and Hamid Laga.
\newblock A comprehensive survey of deep learning for image captioning.
\newblock {\em ACM Computing Surveys (CsUR)}, 51(6):1--36, 2019.

\bibitem{huang2017densely}
Gao Huang, Zhuang Liu, Laurens Van Der~Maaten, and Kilian~Q Weinberger.
\newblock Densely connected convolutional networks.
\newblock In {\em Proceedings of the IEEE Conference on Computer Vision and
  Pattern Recognition}, pages 4700--4708, 2017.

\bibitem{irvin2019chexpert}
Jeremy Irvin, Pranav Rajpurkar, Michael Ko, Yifan Yu, Silviana Ciurea-Ilcus,
  Chris Chute, Henrik Marklund, Behzad Haghgoo, Robyn Ball, Katie Shpanskaya,
  et~al.
\newblock Chexpert: A large chest radiograph dataset with uncertainty labels
  and expert comparison.
\newblock In {\em Proceedings of the AAAI conference on artificial
  intelligence}, volume~33, pages 590--597, 2019.

\bibitem{jain2021radgraph}
Saahil Jain, Ashwin Agrawal, Adriel Saporta, Steven Truong, Tan Bui, Pierre
  Chambon, Yuhao Zhang, Matthew~P Lungren, Andrew~Y Ng, Curtis Langlotz, et~al.
\newblock Radgraph: Extracting clinical entities and relations from radiology
  reports.
\newblock In {\em Thirty-fifth Conference on Neural Information Processing
  Systems Datasets and Benchmarks Track (Round 1)}, 2021.

\bibitem{Johnson2019MIMICCXRAL}
Alistair E.~W. Johnson, Tom~J. Pollard, Seth~J. Berkowitz, Nathaniel~R.
  Greenbaum, Matthew~P. Lungren, Chih ying Deng, Roger~G. Mark, and Steven
  Horng.
\newblock Mimic-cxr: A large publicly available database of labeled chest
  radiographs.
\newblock {\em arXiv preprint arXiv:1901.07042}, 2019.

\bibitem{Li2018HybridRR}
Christy~Y. Li, Xiaodan Liang, Zhiting Hu, and Eric~P. Xing.
\newblock Hybrid retrieval-generation reinforced agent for medical image report
  generation.
\newblock In {\em Proceedings of the Conference on Neural Information
  Processing Systems}, 2018.

\bibitem{li2019knowledge}
Christy~Y Li, Xiaodan Liang, Zhiting Hu, and Eric~P Xing.
\newblock Knowledge-driven encode, retrieve, paraphrase for medical image
  report generation.
\newblock In {\em Proceedings of the AAAI Conference on Artificial
  Intelligence}, volume~33, pages 6666--6673, 2019.

\bibitem{li2022blip}
Junnan Li, Dongxu Li, Caiming Xiong, and Steven Hoi.
\newblock Blip: Bootstrapping language-image pre-training for unified
  vision-language understanding and generation.
\newblock {\em arXiv preprint arXiv:2201.12086}, 2022.

\bibitem{li2021align}
Junnan Li, Ramprasaath Selvaraju, Akhilesh Gotmare, Shafiq Joty, Caiming Xiong,
  and Steven Chu~Hong Hoi.
\newblock Align before fuse: Vision and language representation learning with
  momentum distillation.
\newblock {\em Advances in neural information processing systems},
  34:9694--9705, 2021.

\bibitem{li2021ffa}
Mingjie Li, Wenjia Cai, Rui Liu, Yuetian Weng, Xiaoyun Zhao, Cong Wang, Xin
  Chen, Zhong Liu, Caineng Pan, Mengke Li, et~al.
\newblock Ffa-ir: Towards an explainable and reliable medical report generation
  benchmark.
\newblock In {\em Thirty-fifth Conference on Neural Information Processing
  Systems Datasets and Benchmarks Track (Round 2)}, 2021.

\bibitem{li2022cross}
Mingjie Li, Wenjia Cai, Karin Verspoor, Shirui Pan, Xiaodan Liang, and Xiaojun
  Chang.
\newblock Cross-modal clinical graph transformer for ophthalmic report
  generation.
\newblock In {\em Proceedings of the IEEE Conference on Computer Vision and
  Pattern Recognition}, pages 20656--20665, 2022.

\bibitem{li2022video}
Mingjie Li, Po-Yao Huang, Xiaojun Chang, Junjie Hu, Yi Yang, and Alex
  Hauptmann.
\newblock Video pivoting unsupervised multi-modal machine translation.
\newblock {\em IEEE Transactions on Pattern Analysis and Machine Intelligence},
  2022.

\bibitem{li2022auxiliary}
Mingjie Li, Rui Liu, Fuyu Wang, Xiaojun Chang, and Xiaodan Liang.
\newblock Auxiliary signal-guided knowledge encoder-decoder for medical report
  generation.
\newblock {\em World Wide Web}, pages 1--18, 2022.

\bibitem{lin-2004-rouge}
Chin-Yew Lin.
\newblock {ROUGE}: A package for automatic evaluation of summaries.
\newblock In {\em Text Summarization Branches Out}. Association for
  Computational Linguistics, July 2004.

\bibitem{lin2014microsoft}
Tsung-Yi Lin, Michael Maire, Serge Belongie, James Hays, Pietro Perona, Deva
  Ramanan, Piotr Doll{\'a}r, and C~Lawrence Zitnick.
\newblock Microsoft coco: Common objects in context.
\newblock In {\em Proceedings of the European Conference on Computer Vision},
  pages 740--755, 2014.

\bibitem{liu2022competence}
Fenglin Liu, Shen Ge, and Xian Wu.
\newblock Competence-based multimodal curriculum learning for medical report
  generation.
\newblock {\em arXiv preprint arXiv:2206.14579}, 2022.

\bibitem{liu2021exploring}
Fenglin Liu, Xian Wu, Shen Ge, Wei Fan, and Yuexian Zou.
\newblock Exploring and distilling posterior and prior knowledge for radiology
  report generation.
\newblock In {\em Proceedings of the IEEE Conference on Computer Vision and
  Pattern Recognition}, pages 13753--13762, 2021.

\bibitem{liu2021contrastive}
Fenglin Liu, Changchang Yin, Xian Wu, Shen Ge, Ping Zhang, and Xu Sun.
\newblock Contrastive attention for automatic chest x-ray report generation.
\newblock In {\em Findings of the Association for Computational Linguistics},
  pages 269--280, 2021.

\bibitem{liu2021auto}
Fenglin Liu, Chenyu You, Xian Wu, Shen Ge, Xu Sun, et~al.
\newblock Auto-encoding knowledge graph for unsupervised medical report
  generation.
\newblock In {\em Proceedings of the Conference on Neural Information
  Processing Systems}, pages 16266--16279, 2021.

\bibitem{liu2020k}
Weijie Liu, Peng Zhou, Zhe Zhao, Zhiruo Wang, Qi Ju, Haotang Deng, and Ping
  Wang.
\newblock K-bert: Enabling language representation with knowledge graph.
\newblock In {\em Proceedings of the AAAI Conference on Artificial
  Intelligence}, volume~34, pages 2901--2908, 2020.

\bibitem{loshchilov2017decoupled}
Ilya Loshchilov and Frank Hutter.
\newblock Decoupled weight decay regularization.
\newblock {\em arXiv preprint arXiv:1711.05101}, 2017.

\bibitem{nooralahzadeh2021progressive}
Farhad Nooralahzadeh, Nicolas~Perez Gonzalez, Thomas Frauenfelder, Koji
  Fujimoto, and Michael Krauthammer.
\newblock Progressive transformer-based generation of radiology reports.
\newblock {\em arXiv preprint arXiv:2102.09777}, 2021.

\bibitem{papineni-etal-2002-bleu}
Kishore Papineni, Salim Roukos, Todd Ward, and Wei-Jing Zhu.
\newblock {B}leu: a method for automatic evaluation of machine translation.
\newblock In {\em Proceedings of the 40th Annual Meeting of the Association for
  Computational Linguistics}, July 2002.

\bibitem{shin2016learning}
Hoo-Chang Shin, Kirk Roberts, Le Lu, Dina Demner-Fushman, Jianhua Yao, and
  Ronald~M Summers.
\newblock Learning to read chest x-rays: Recurrent neural cascade model for
  automated image annotation.
\newblock In {\em Proceedings of the IEEE Conference on Computer Vision and
  Pattern Recognition}, pages 2497--2506, 2016.

\bibitem{song2022cross}
Xiao Song, Xiaodan Zhang, Junzhong Ji, Ying Liu, and Pengxu Wei.
\newblock Cross-modal contrastive attention model for medical report
  generation.
\newblock In {\em Proceedings of the 29th International Conference on
  Computational Linguistics}, pages 2388--2397, 2022.

\bibitem{vaswani2017attention}
Ashish Vaswani, Noam Shazeer, Niki Parmar, Jakob Uszkoreit, Llion Jones,
  Aidan~N Gomez, {\L}ukasz Kaiser, and Illia Polosukhin.
\newblock Attention is all you need.
\newblock In {\em Proceedings of the Conference on Neural Information
  Processing Systems}, 2017.

\bibitem{vedantam2015cider}
Ramakrishna Vedantam, C Lawrence~Zitnick, and Devi Parikh.
\newblock Cider: Consensus-based image description evaluation.
\newblock In {\em Proceedings of the IEEE Conference on Computer Vision and
  Pattern Recognition}, 2015.

\bibitem{wuchest}
Joy Wu, Nkechinyere Agu, Ismini Lourentzou, Arjun Sharma, Joseph Paguio,
  Jasper~Seth Yao, Edward~Christopher Dee, William Mitchell, Satyananda
  Kashyap, Andrea Giovannini, et~al.
\newblock Chest imagenome dataset.
\newblock {\em PhysioNet}, 2021.

\bibitem{wu2022multimodal}
Xing Wu, Jingwen Li, Jianjia Wang, and Quan Qian.
\newblock Multimodal contrastive learning for radiology report generation.
\newblock {\em Journal of Ambient Intelligence and Humanized Computing}, pages
  1--10, 2022.

\bibitem{yan2021weakly}
An Yan, Zexue He, Xing Lu, Jiang Du, Eric Chang, Amilcare Gentili, Julian
  McAuley, and Chun-nan Hsu.
\newblock Weakly supervised contrastive learning for chest x-ray report
  generation.
\newblock In {\em Findings of the Association for Computational Linguistics:
  EMNLP 2021}, pages 4009--4015, 2021.

\bibitem{yang2022knowledge}
Shuxin Yang, Xian Wu, Shen Ge, S~Kevin Zhou, and Li Xiao.
\newblock Knowledge matters: Chest radiology report generation with general and
  specific knowledge.
\newblock {\em Medical Image Analysis}, page 102510, 2022.

\bibitem{zhang2020contrastive}
Yuhao Zhang, Hang Jiang, Yasuhide Miura, Christopher~D Manning, and Curtis~P
  Langlotz.
\newblock Contrastive learning of medical visual representations from paired
  images and text.
\newblock {\em arXiv preprint arXiv:2010.00747}, 2020.

\bibitem{zhang2020radiology}
Yixiao Zhang, Xiaosong Wang, Ziyue Xu, Qihang Yu, Alan Yuille, and Daguang Xu.
\newblock When radiology report generation meets knowledge graph.
\newblock In {\em Proceedings of the AAAI Conference on Artificial
  Intelligence}, volume~34, pages 12910--12917, 2020.

\bibitem{zhang2021biomedical}
Yuhao Zhang, Yuhui Zhang, Peng Qi, Christopher~D Manning, and Curtis~P
  Langlotz.
\newblock Biomedical and clinical english model packages for the stanza python
  nlp library.
\newblock {\em Journal of the American Medical Informatics Association},
  28(9):1892--1899, 2021.

\bibitem{zhao2021automatic}
Haifeng Zhao, Jie Chen, Lili Huang, Tingting Yang, Wanhai Ding, and Chuanfu Li.
\newblock Automatic generation of medical report with knowledge graph.
\newblock In {\em Proceedings of the International Conference on Computing and
  Pattern Recognition}, pages 1--1, 2021.

\bibitem{zheng2022rethinking}
Yizhen Zheng, Shirui Pan, Vincent~Cs Lee, Yu Zheng, and Philip~S Yu.
\newblock Rethinking and scaling up graph contrastive learning: An extremely
  efficient approach with group discrimination.
\newblock In {\em Proceedings of the Conference on Neural Information
  Processing Systems}, 2022.

\end{thebibliography}
}

\end{document}